\documentclass[11pt]{article}

\usepackage[final]{acl}

\usepackage{times}
\usepackage{latexsym}

\usepackage[T1]{fontenc}

\usepackage[utf8]{inputenc}

\usepackage{microtype}
\usepackage{amsmath}
\usepackage{inconsolata}

\usepackage{graphicx}

\usepackage{hyphenat}
\usepackage{booktabs,tabularx,longtable,siunitx, multirow}
\usepackage{array}

\usepackage{listings}
\usepackage[most]{tcolorbox}
\tcbuselibrary{breakable,listings}

\usepackage{enumitem}

\lstdefinestyle{promptstyle}{
  basicstyle=\ttfamily\small,
  breaklines=true,
  columns=fullflexible,
  keepspaces=true,
  showstringspaces=false
}

\newtcolorbox{promptbox}{
  colback=gray!4, colframe=gray!60,
  boxrule=0.4pt, arc=2pt,
  left=6pt, right=6pt, top=6pt, bottom=6pt,
  enhanced, breakable   
}

\title{Empowering LLM Agents with Geospatial Awareness:\\
Toward Grounded Reasoning for Wildfire Response}

\author{
Yiheng Chen$^{1}$, \quad
Lingyao Li$^{2}$\thanks{Corresponding authors.}, \quad
Zihui Ma$^{3}$, \quad
Qikai Hu$^{4}$, \\
\textbf{Yilun Zhu$^{4}$, \quad
Min Deng$^{5}$, \quad
Runlong Yu$^{1}$\footnotemark[1]}\\
$^{1}$University of Alabama \quad
$^{2}$University of South Florida \quad
$^{3}$New York University\\
$^{4}$University of Michigan \quad
$^{5}$Texas Tech University\\
\texttt{ychen226@crimson.ua.edu}, 
\texttt{lingyaol@usf.edu}, 
\texttt{zm2864@nyu.edu},\\
\texttt{\{wsxgshqk,allanzhu\}@umich.edu},
\texttt{mindeng@ttu.edu}, 
\texttt{ryu5@ua.edu}
}

\begin{document}
\maketitle
\begin{abstract}
Effective disaster response is essential for safeguarding lives and property. Existing statistical approaches often lack semantic context, generalize poorly across events, and offer limited interpretability. 
While Large language models (LLMs) provide few-shot generalization, they remain text-bound and blind to geography. 
To bridge this gap, we introduce a \textbf{Geospatial Awareness Layer (GAL)} that grounds LLM agents in structured earth data. 
Starting from raw wildfire detections, 
GAL automatically retrieves and integrates infrastructure, demographic, terrain, and weather information from external geodatabases, assembling them into a concise, unit-annotated \emph{perception script}.
This enriched context enables agents to produce evidence-based resource-allocation recommendations (e.g., personnel assignments, budget allocations), further reinforced by historical analogs and daily change signals for incremental updates. We evaluate the framework in real wildfire scenarios across multiple LLM models, showing that geospatially grounded agents can outperform baselines. 
The proposed framework can generalize to other hazards such as floods and hurricanes. 
\href{https://github.com/defene/GAL}{Code Availability.}
\end{abstract}


\section{Introduction}

Natural hazards like wildfires, floods, earthquakes, and hurricanes evolve rapidly, imposing intense time and space constraints on the disaster decision-making process~\citep{ma2024investigating,li2023exploring}. Wildfires in particular illustrate these challenges: they can spread quickly across vast areas, with propagation behavior influenced by complex topography and weather characteristics~\citep{finney1998farsite,Abatzoglou2016PNAS}. Effective disaster response is critical for protecting lives, property, and enhancing community resilience under these extreme conditions~\citep{Cutter2016Resilience}. However, emergency managers often face the difficulty of estimating personnel, equipment, and financial resources across diverse regions due to incomplete and uncertain information. While existing approaches in wildfire contexts, such as hotspot counts, fire radiative power (FRP) statistics, and time-series predictors, provide useful signals, they often ignore context and lack generalizability and interpretability~\citep{Giglio2016MODIS,Wooster2005FRP}. These limitations underscore an urgent need for effective disaster decision-support systems, especially in wildfire settings where operational decisions are made under both environmental uncertainty and rapidly shifting information conditions~\citep{Dunn2017Risk,Thompson2019RiskAnalytics,Martell2015FMDSS,chen2026politicized}.

Large language models (LLMs) offer a promising direction, with strong few-shot generalization and the capacity to integrate diverse sources of knowledge~\citep{Brown2020GPT3,Touvron2023LLaMA2,lei2025harnessing}. More broadly, recent work on foundation models for environmental science has highlighted their potential to unify heterogeneous Earth data and support downstream scientific and decision-making tasks~\citep{yu2025survey}. Despite this potential, current LLMs remain weak in geospatial grounding. They are prone to hallucinating in numerical operations and often struggle to reliably localize, interpret, or reason over the spatial context that is crucial for appraising hazard impacts and response needs~\citep{Huang2023HallucinationLLM,li2025pixels}. This disconnect between linguistic adaptability and real-world awareness represents a central limitation of current LLM agents. Inspired by recent advances in augmenting LLMs with external knowledge, together with the emerging need for retrieval mechanisms tailored to Earth and environmental sciences~\citep{lewis2020retrieval,Borgeaud2022RETRO,yu2025rag}, we pose an important question: can LLM agents draw on structured geospatial data from our physical world to support disaster response decisions?

Turning this idea requires bridging LLMs with the physical world along several dimensions. First, the agent needs to actively retrieve relevant geospatial evidence at appropriate spatial and temporal scales. Second, the retrieved information has to be represented in a form that LLMs can interpret without being overwhelmed by raw geometries or large tabular data. Third, the system should generate stable and actionable outputs that remain consistent across days and events.

We address these challenges with a novel framework that equips LLM agents with a \textbf{Geospatial Awareness Layer (GAL)}, providing a structured interface to access data from the physical world. First, for retrieval, we build a PostGIS–raster database that~\citep{obe2021postgis}, given only hotspot coordinates and timestamps, automatically extracts infrastructure, demographic, terrain, and weather attributes using spatial joins and zonal statistics. Second, for representation, the layer encodes these heterogeneous signals into a compact, unit-annotated perception script with fixed fields, abstracting away raw geometries and large tables into a stable form that LLMs can readily reason over. Third, for decision stability, it enforces schema validation, unit normalization, and bounded ranges, while also reinforcing outputs with historical analogs and daily change signals.
Our key contributions include: 
\begin{itemize}[nosep]
\item \textbf{Geospatially grounded framework.} We introduce a novel framework that enables LLM agents to see and interpret the physical world. This layer actively retrieves and interprets spatial evidence to ground decision-making in real-world conditions. 
\item \textbf{Reasoning analysis on real wildfires.} 
We evaluate grounded LLMs on multiple California wildfire events to analyze how spatial grounding affects reasoning behavior, finding improved interpretability and alignment with operational outcomes. 
\item \textbf{Open benchmark.} We conduct a cross-model evaluation of LLMs within this framework, establishing one of the first benchmarks for assessing geospatially grounded reasoning and decision accuracy in wildfire response.
\end{itemize}

\section{Related Work}

\subsection{Traditional Wildfire Modeling}

Traditional wildfire modeling spans physical simulation, statistical analysis, and remote sensing. Physical fire spread models, such as the Rothermel equation~\citep{rothermel1972mathematical} and its implementation in FARSITE~\citep{finney1998farsite}, provide mechanistic predictions of fire growth. Coupled fire–atmosphere systems like WRF-Fire~\citep{coen2013wrf} further integrate weather dynamics (e.g., wind, humidity) with combustion processes. Statistical and time-series approaches estimate wildfire occurrence or extent from historical patterns~\citep{preisler2007statistical, laube2021wildfire}, other data-driven methods predict burned area~\citep{cortez2007} or assess fire impacts through epidemic-style models~\citep{ma2024investigating}. Remote sensing techniques complement these efforts by detecting and forecasting wildfire activity using satellite observations~\citep{chaparro2016predicting, huot2022next}. Despite their value for scientific understanding, these approaches are often slow and lack contextual awareness. 

\begin{figure*}[t]
    \centering
    \includegraphics[width=\textwidth]{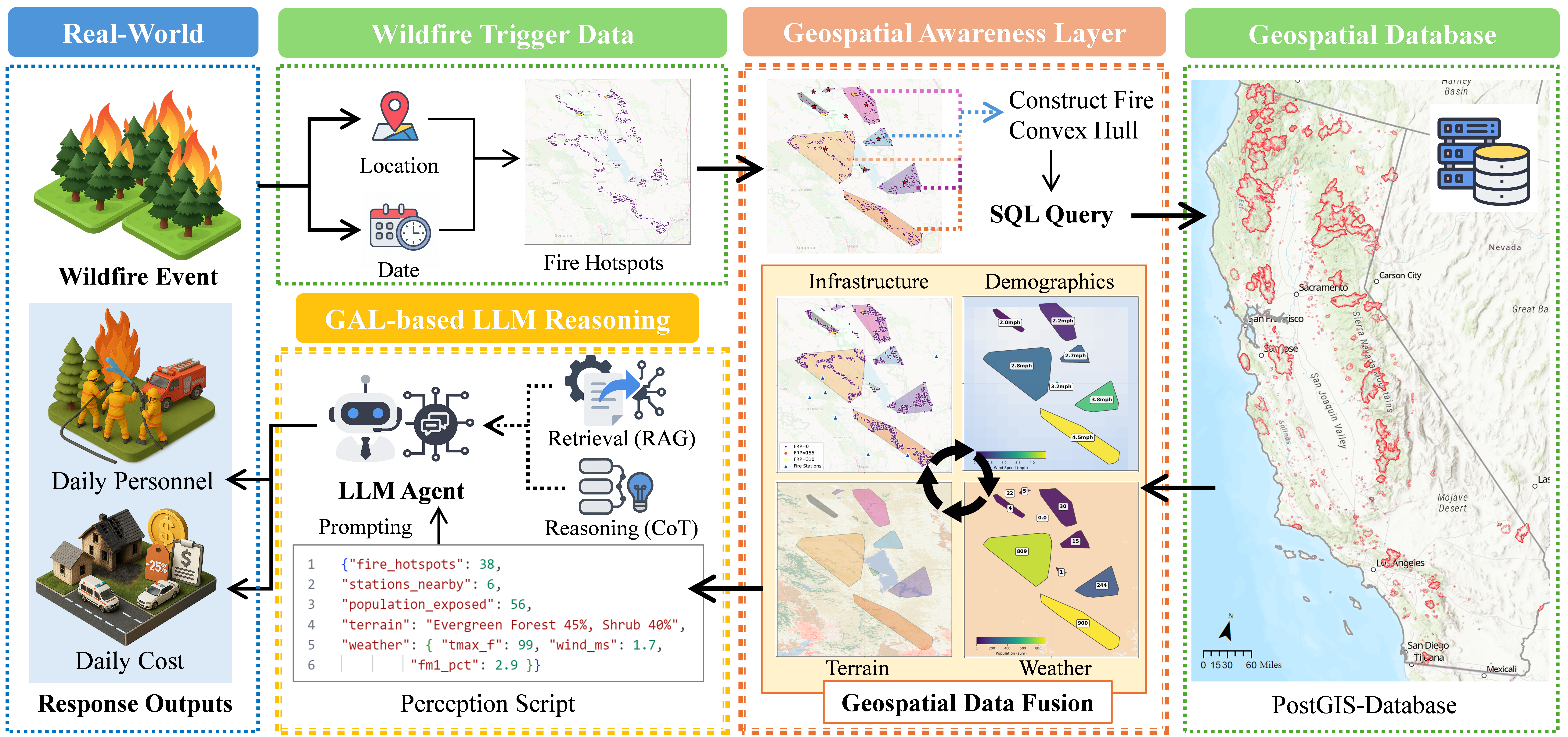}
    \caption{An illustration of the research framework.}
    \label{fig:framework}
\end{figure*}

\subsection{LLMs for Disaster Response}

Compared with traditional techniques, LLMs offer unique advantages in processing multimodal inputs, analyzing unstructured text, and generating interpretable outputs to support decision-making in dynamic environments~\citep{li2024logicity,li2025pixels,chen2024enhancing}. Recent studies have explored their potential for disaster response~\citep{chen2025perceptions,otal2024llm,goecks2023disasterresponsegpt}. For instance, pretrained LLMs have been used to extract location-based descriptors from social media, showing that incorporating geographic knowledge can improve performance~\citep{hu2023geo}. 
Other works integrate maps and geospatial data into LLM workflows to enhance disaster-relevant analysis~\citep{hu2023geo,yin2025llm,zguir2025roadmind}. Despite these advances, most LLMs operate in text space and cannot reliably perceive or reason about the geospatial conditions that govern real-world hazards~\citep{Huang2023HallucinationLLM,li2025pixels}.

LLM-based approaches to disaster prediction and simulation are particularly relevant to our focus~\citep{tang2025predicting,li2025llms,chen2025perceptions,zhong2024hmp}. For example, \citet{li2025llms} evaluate LLMs as world models to simulate human-perceived seismic risk using seismic data and Google Street views, revealing geospatial patterns consistent with ``Did You Feel It?'' maps \citep{quitoriano2020usgs}. Within wildfire studies, domain-specific models such as WildfireGPT~\citep{xie2024wildfiregpt} have been introduced for response tasks, and later evaluated for quantitative fire spread prediction~\citep{ramesh2025assessing}. It has been observed that while general LLMs can describe wildfire behavior qualitatively, they struggle with accurate quantitative forecasts (e.g., fire radiative power). LLMs have also been used for evacuation prediction~\citep{dang2025large,chen2025perceptions}; for instance, \citet{chen2025perceptions} develop a framework that outperforms traditional theory-based models in modeling evacuation behaviors.

Our work differs from prior studies by introducing a GAL that dynamically grounds LLM agents in structured Earth data. 
GAL can perform inference-time retrieval of geospatial attributes from live databases and organizes them into a stable, unit-annotated representation. This dynamic grounding enables LLMs to produce geographically coherent and operationally actionable outputs across evolving disaster contexts.

\section{Methodology}

\subsection{Framework Design}

We propose to shift language models from closed linguistic systems toward agents capable of perceiving and reasoning about the physical world.
Existing LLM-based tools remain confined to text, being powerful in text reasoning and summarization, yet blind to geography. 
To bridge this gap, we introduce a \textbf{Geospatial Awareness Layer (GAL)} that serves as an environmental sensor that provides active retrieval and integration of spatial evidence.

As illustrated in Figure~\ref{fig:framework}, the proposed framework links real wildfire events to actionable response decisions through geospatially grounded reasoning. Each event is characterized by its detection date and hotspot locations, which trigger our GAL. GAL clusters hotspots into fire footprints, constructs convex hulls, and issues SQL queries to retrieve spatial attributes—such as infrastructure, demographics, terrain, and weather—from a PostGIS–raster database. These layers are fused into a compact, unit-annotated perception script that summarizes local exposure, accessibility, and environmental conditions. The script then drives the GAL-based LLM Reasoning module, where the model performs retrieval-augmented and chain-of-thought reasoning over current states and historical analogs to produce evidence-grounded recommendations for daily personnel and cost.

\subsection{Data Preparation}
\label{sec:Data_Preparation}
We integrate multiple open geospatial datasets, including fire detection, infrastructure, demographics, terrain, and weather.  
For event initialization, we use satellite-derived wildfire detections from the Visible Infrared Imaging Radiometer Suite (VIIRS) sensors onboard the Suomi National Polar-orbiting Partnership (S-NPP), NOAA-20, and NOAA-21 satellites. The VIIRS active fire data, with a spatial resolution of 375m, are obtained from NASA’s Fire Information for Resource Management System (FIRMS)~\citep{FIRMS}.  
Administrative and population attributes are derived from county boundaries and demographic statistics provided by the U.S. Census Bureau’s American Community Survey (ACS) and the CA-POP dataset~\citep{capop}.  
Critical facilities such as fire stations and infrastructure are obtained from the Homeland Infrastructure Foundation-Level Data (HIFLD) repository.  
Land surface information is provided by the National Land Cover Database (NLCD)~\citep{NLCD2021} and a digital elevation model (DEM) for terrain metrics.  
Daily weather variables, including Burning Index (BI; a composite fire-danger metric used by agencies to gauge escalation risk), temperature, wind speed, and 1-hour fuel moisture (FM1; moisture content of fine fuels that governs ignition and spread rate), are sourced from the gridMET (METDATA) dataset~\citep{WEATHER}.  
For evaluation, we further process daily situation reports from ICS-209-Plus~\citep{ics209plus} (standardized daily logs of personnel assignments, cumulative costs, and containment status filed by incident command), extracting daily personnel and cost records. These records are cleaned and aggregated into per-event time series for ground-truth validation. 

The fundamental unit of analysis is a \emph{fire--day}: a single wildfire incident on a single calendar date, paired with one perception script and one ground-truth target pair (daily personnel, daily cost from ICS-209-Plus). Dynamic inputs—VIIRS hotspot statistics, gridMET weather variables, and the daily fire footprint—are all synchronized to the same calendar date, while static layers (NLCD land cover, ACS demographics, HIFLD infrastructure) are joined to that day's footprint via spatial overlays and zonal statistics, yielding a consistent spatio-temporal tile per instance. Fire--days with missing or inconsistent ICS-209-Plus records are excluded from training and evaluation without imputation. For geospatial attributes, genuinely absent fields are flagged \texttt{NA} in the perception script rather than filled with zero; short gaps in weather observations are filled using a one-to-two day rolling window to avoid leaking future information.

\subsection{Geospatial Awareness Layer (GAL)}

GAL acts as a geospatial encoder that converts raw spatial signals into interpretable symbolic representations, allowing the LLM to reason over structured context rather than unbounded numeric inputs. By linking real-world coordinates to semantically meaningful fields, GAL provides the spatial grounding that conventional LLMs lack.

\paragraph{Event detection and normalization.}
GAL takes as input the detection date and hotspot coordinates with basic attributes. It clusters hotspots using DBSCAN~\cite{Ester1996DBSCAN} (radius = 3 km, MinPts $\geq$ 3). When fewer points are available, the module builds simple geometric buffers—a line for two points or a small circle for one—to preserve spatial continuity. For each cluster, FRP-weighted centroids are calculated, and polygonal footprints are generated using convex hulls or buffers to outline the fire’s active area. The resulting standardized event footprints are stored in PostGIS with daily timestamps for subsequent spatial queries.

\paragraph{Geospatial information retrieval.}
Given the normalized fire footprints, GAL acts as an automatic retrieval interface that converts spatial context into semantically structured attributes usable by the LLM. 
It composes parameterized SQL queries over a PostGIS–raster store to gather relevant evidence while preserving spatial and temporal alignment. Each query retrieves a specific dimension of the wildfire environment: (i) \textbf{Fire hotspots \& stations:} nearest-$k$ fire stations (geodesic distance, $k=3$), station density within radius, and intra-day hotspot counts/FRP per cluster. (ii) \textbf{Demographics (exposure):} population sum/density within each footprint and county identifiers for cross-jurisdiction reasoning. (iii) \textbf{Terrain (operational complexity):} NLCD land-cover composition per cluster, Shannon diversity and fragmentation indices, classification of vegetation into high/medium/low spread-risk classes, estimated fractions of continuous fuels and natural/artificial barriers, and an area-weighted spread-potential score. (iv) \textbf{Weather (escalation risk):} FRP-weighted Burning Index, wind speed, air temperature (Tmax/Tmin), and 1-hr fuel moisture (FM1) aggregated over the footprint and aligned to the analysis day.
Scale adapts per query (cluster footprint vs.\ county buffer) and per variable (daily vs.\ hourly products). All joins and raster stats are executed server-side to minimize token and bandwidth costs.

\begin{figure*}[t]
    \centering
    \includegraphics[width=0.82\textwidth]{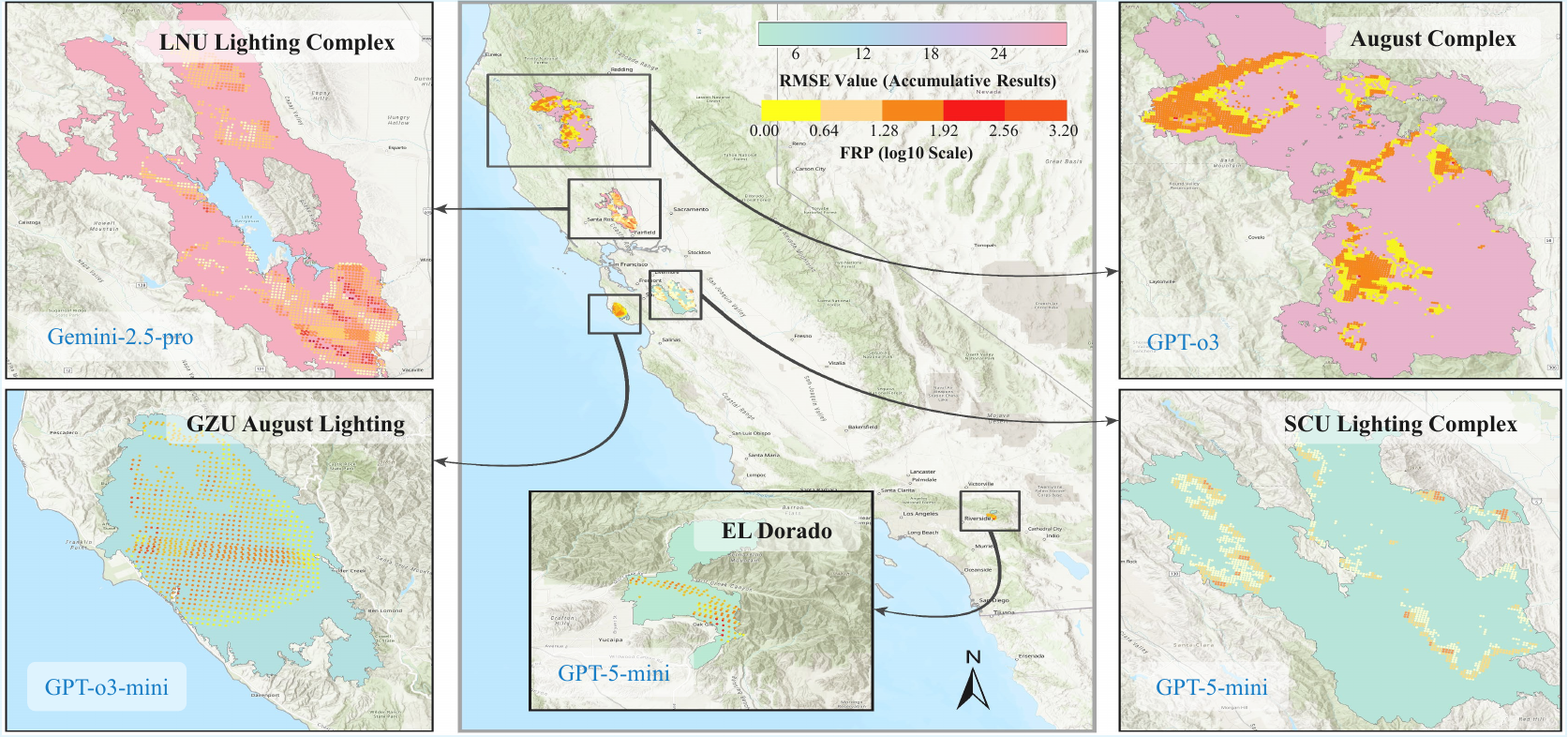}
    \caption{Predictive accuracy across five 2020 California wildfires and hotspot distributions colored by FRP.}
    \label{fig:fig2}
\end{figure*}

\paragraph{Feature consolidation.}
GAL consolidates retrieved signals into cluster-level features and a global snapshot. 
Cluster features include point count, sum FRP, max brightness, hotspot-weighted weather, land-cover mix, exposed population, and nearest-station distances. 
The global snapshot contains totals (points/FRP), extremes (max FRP, max brightness), and robust distributional measures (e.g., median FRP-per-cluster, $p95$). 
For $t>1$, GAL computes temporal anchors: 3-day/7-day rolling means and maxima, as well as qualitative deltas (``$\uparrow$'' or ``$\approx$'' or ``$\downarrow$'') for points, FRP, burned area, and key weather fields. 
These stabilize incremental decisions without requiring cluster lineage.

\subsection{GAL-based LLM Reasoning}

We transform geospatial information into a structured, language-ready context that guides LLMs through grounded reasoning. The process combines three elements: a permutation-invariant, unit-locked perception script that standardizes spatial inputs; a retrieval-augmented module that supplies historical analogs as priors; and a rubric-guided chain-of-thought (CoT) that organizes reasoning into interpretable and physically consistent steps.

\paragraph{Perception script.}
GAL renders heterogeneous geospatial evidence into a compact, unit-annotated perception script with fixed slots and permutation invariance, so the agent reasons over stable, semantically aligned inputs rather than raw geometries or long tables. The script contains: (i) \textbf{Top-$K$ cluster summaries} for the most consequential activity areas; (ii) a \textbf{global snapshot} of the event day; and (iii) a \textbf{unit lock/normalization} step (temperatures in Kelvin, wind in m/s, FM1 in \%, standardized magnitudes). Missing fields are marked \texttt{NA} with defaults to reduce hallucination and numeric drift.

\paragraph{Analogs scaffold.}
To supply scale priors, we attach a shared historical-analogs module based on retrieval-augmented generation~\citep{lewis2020retrieval,yu2025rag}. This module represents each event--day snapshot as a standardized numeric vector comprising z-scored features for fire activity, exposure, terrain, and weather, along with categorical flags. Weighted cosine similarity retrieves the top-$k$ prior analog days with unique-by-fire de-duplication. For active fire days, the vector is built from current GAL signals; for no-hotspot days, a rolling 3/7-day trajectory vector is used instead. These ranges provide soft bounds for personnel and cost estimation.

\paragraph{Prompted reasoning with rubric-guided CoT.}
We transform the script and analog priors into an auditable decision via a rubric-guided CoT~\citep{Wei2022CoT}. The prompt requires brief justifications on terrain, weather, fire intensity, exposure, and access, and produces a structured JSON with six intermediate indicators \{spread difficulty, deployment access, weather risk, terrain complexity, exposure density, station coverage\} on a five-level scale. Two modes share one interface: \emph{Day-1} reasons from the current script plus top-$k$ analogs; \emph{Incremental} additionally consumes yesterday’s outputs and compact change cues, prompting a short comparative rationale before adjustment. A validator enforces JSON schema, unit correctness, and bounded ranges; failures trigger targeted re-prompts. Stability comes from fixed field slots, normalized magnitudes, and analog-anchored bounds. Example prompts are provided in \textbf{Appendix~\ref{app:Prompt_Design}}.

\section{Experiments}

\subsection{Dataset}
Following the data sources described in Section~\ref{sec:Data_Preparation}, we evaluate the proposed framework on fourteen large wildfire events that occurred in California during the 2020 season.  
The dataset spans diverse ignition sources, terrains, and durations (see Table~\ref{tab:fire-durations} in \textbf{Appendix~\ref{app:Fire_Statistics}}), covering incidents from late July through early October.  
Among them, five major events (\textsc{CZU Aug Lightning}, \textsc{El Dorado}, \textsc{LNU Lightning Complex}, \textsc{SCU Lightning Complex}, and \textsc{August Complex}) are held out for evaluation, while the remaining nine incidents constitute the historical corpus used by the RAG module and for training non-LLM baselines.  
This split ensures spatial and temporal disjointness between training and evaluation while maintaining realistic wildfire diversity.

\begin{table*}[t]
  \centering
  \footnotesize
  \setlength{\tabcolsep}{6pt}
  \renewcommand{\arraystretch}{1}
  \begin{tabular}{c *{10}{S[table-format=1.4]}}
    \toprule
    \multirow{2}{*}{Models}
      & \multicolumn{2}{c}{CZU}
      & \multicolumn{2}{c}{ELDO}
      & \multicolumn{2}{c}{LNU}
      & \multicolumn{2}{c}{SCU}
      & \multicolumn{2}{c}{AUGC}\\
    \cmidrule(lr){2-3}\cmidrule(lr){4-5}\cmidrule(lr){6-7}\cmidrule(lr){8-9}\cmidrule(lr){10-11}
     & {MAE} & {RMSE} & {MAE} & {RMSE} & {MAE} & {RMSE} & {MAE} & {RMSE} & {MAE} & {RMSE} \\
    \midrule
    Physical & 3.2178 & 3.6681 & 1.2623 & 1.4817 & 4.4987 & 4.9380 & 1.9013 & 2.1603 & 2.6976 & 3.6299 \\
    LSTM     & 5.5422 & 5.9243 & 2.7959 & 3.1506 & 6.8444 & 7.1553 & 3.3554 & 3.8586 & 4.3175 & 4.9412 \\
    \midrule
    Gemini\hyp{}2.5\hyp{}Flash & 4.4517 & 5.1462 & 1.0977 & 1.3941 & 4.6493 & 5.3697 & 2.0380 & 2.8660 & 3.2682 & 3.9792 \\
    GPT\hyp{}4o\hyp{}mini      & 4.0891 & 4.6469 & 1.9234 & 2.1657 & 3.9590 & 4.5367 & {\bfseries 1.0770} & {\bfseries 1.5466} & 3.8893 & 4.3974 \\
    GPT\hyp{}4o                & 4.0409 & 4.6471 & 1.2134 & 1.4672 & 3.6717 & 4.3117 & 1.8040 & 2.4694 & 3.2277 & 4.0495 \\
    GPT\hyp{}4.1\hyp{}mini     & 3.9966 & 4.6667 & 1.5977 & 1.9122 & 4.5579 & 5.2680 & 1.8740 & 2.8371 & 2.6463 & 3.3665 \\
    GPT\hyp{}4.1               & 4.0974 & 4.8116 & 1.7389 & 1.9735 & 4.3107 & 5.0133 & 2.7487 & 3.9777 & 3.6742 & 4.5958 \\
    GPT\hyp{}5\hyp{}mini       & \underline{2.4600} & \underline{3.0067} & {\bfseries 0.9086} & {\bfseries 1.0845} & \underline{2.6693} & \underline{3.1717} & 3.0500 & 3.3324 & \underline{2.5408} & {\bfseries 3.3326} \\
    GPT\hyp{}5                 & 2.8482 & 3.2180 & \underline{0.9126} & \underline{1.1722} & 3.0859 & 3.7891 & 1.5530 & 1.7762 & 2.8172 & 3.6462 \\
    \midrule
    Gemini\hyp{}2.5\hyp{}Pro   & 4.8717 & 5.7561 & 1.9129 & 2.4405 & 4.1486 & 4.9068 & 3.1723 & 4.7006 & 3.2702 & 3.9927 \\
    GPT\hyp{}o3\hyp{}mini      & {\bfseries 2.2826} & {\bfseries 2.8561} & 1.6263 & 1.8525 & {\bfseries 1.2259} & {\bfseries 1.6785} & 1.5093 & 1.6869 & {\bfseries 2.5165} & \underline{3.3576} \\
    GPT\hyp{}o3                & 3.2583 & 3.7656 & 1.2737 & 1.5457 & 4.0952 & 4.7052 & \underline{1.3927} & \underline{1.6506} & 2.7599 & 3.4331 \\
    \bottomrule
  \end{tabular}
  \vspace{2pt}
  {\scriptsize Abbrev.: CZU = CZU Aug. Lightning; ELDO = El Dorado; LNU = LNU Lightning; SCU = SCU Lightning; AUGC = August Complex.}
    \caption{Daily personnel prediction results (MAE/RMSE) on five held-out wildfires.}
    \label{tab:personnel}
\end{table*}

\begin{table*}[t]
  \centering
    \footnotesize
  \setlength{\tabcolsep}{6pt}
  \renewcommand{\arraystretch}{1}
  \begin{tabular}{c *{10}{S[table-format=1.4]}}
    \toprule
    \multirow{2}{*}{Models}
      & \multicolumn{2}{c}{CZU}
      & \multicolumn{2}{c}{ELDO}
      & \multicolumn{2}{c}{LNU}
      & \multicolumn{2}{c}{SCU}
      & \multicolumn{2}{c}{AUGC}\\
    \cmidrule(lr){2-3}\cmidrule(lr){4-5}\cmidrule(lr){6-7}\cmidrule(lr){8-9}\cmidrule(lr){10-11}
     & {MAE} & {RMSE} & {MAE} & {RMSE} & {MAE} & {RMSE} & {MAE} & {RMSE} & {MAE} & {RMSE} \\
    \midrule
    Physical        & 1.2484 & 1.3979 & 0.9421 & 1.4042 & 2.9998 & 3.5513 & 1.6222 & 1.9078 & 1.9755 & 2.5039 \\
    LSTM            & 1.2009 & 1.7702 & 1.3117 & 1.9195 & 3.7605 & 4.4340 & 2.2523 & 2.7913 & 1.8712 & 2.6876 \\
    \midrule
    Gemini\hyp{}2.5\hyp{}flash & 1.4029 & 1.7572 & {\bfseries 0.7971} & 1.3868 & 2.7750 & 3.5688 & 1.6066 & 1.9672 & 2.2081 & 2.6126 \\
    GPT\hyp{}4o\hyp{}mini      & 1.3520 & 1.6307 & 0.8006 & 1.4770 & 2.9398 & 3.7167 & \underline{1.2701} & 1.5829 & 2.1567 & 2.5263 \\
    GPT\hyp{}4o                & 1.3071 & 1.5944 & 0.8062 & 1.3992 & \underline{2.4031} & \underline{3.2454} & 1.3549 & 1.7213 & 1.8794 & 2.3381 \\
    GPT\hyp{}4.1\hyp{}mini     & 1.2128 & 1.5345 & 0.7999 & 1.4345 & 3.0051 & 3.8043 & 1.4321 & 1.7667 & 1.8045 & 2.3087 \\
    GPT\hyp{}4.1               & 1.5083 & 1.8928 & 0.8706 & 1.5239 & 2.9922 & 3.8010 & 2.1056 & 2.7367 & 2.0662 & 2.5178 \\
    GPT\hyp{}5\hyp{}mini       & \underline{1.0229} & 1.3740 & \underline{0.7973} & \underline{1.3633} & 2.5250 & {\bfseries 3.1504} & {\bfseries 0.9999} & {\bfseries 1.3888} & 2.0941 & 2.4502 \\
    GPT\hyp{}5                 & 1.0230 & \underline{1.3284} & 0.8072 & 1.4238 & 2.6781 & 3.3963 & 1.2822 & \underline{1.5542} & \underline{1.7897} & {\bfseries 2.2270} \\
    \midrule
    Gemini\hyp{}2.5\hyp{}pro   & 1.5842 & 2.1840 & 1.3254 & 2.0212 & {\bfseries 2.3426} & 3.3043 & 2.0097 & 2.5964 & 2.2865 & 2.6452 \\
    GPT\hyp{}o3\hyp{}mini      & {\bfseries 0.9530} & {\bfseries 1.2599} & 0.8418 & {\bfseries 1.3535} & 2.7487 & 3.3869 & 1.6596 & 1.9789 & 1.8401 & 2.2779 \\
    GPT\hyp{}o3                & 1.2494 & 1.5501 & 0.8804 & 1.5253 & 2.9812 & 3.7631 & 1.7996 & 2.1649 & {\bfseries 1.7874} & \underline{2.2328} \\
    \bottomrule
  \end{tabular}
  \vspace{2pt}
  {\scriptsize Abbrev.: CZU = CZU Aug. Lightning; ELDO = El Dorado; LNU = LNU Lightning; SCU = SCU Lightning; AUGC = August Complex.}
  \caption{Daily cost prediction results (MAE/RMSE) on five held-out wildfires.}
   \label{tab:cost}
\end{table*}

\subsection{Benchmark Models}
To establish a unified benchmark for wildfire response, our evaluation consists of the following:

\noindent\textbf{Baselines.}
The Physical baseline converts fire radiative power and area into flame length via fireline intensity, assigns an NWCG flame-length class, constructs a workload score (perimeter $\times$ class, with minor cluster adjustments), and linearly calibrates it to the target variables.  
The LSTM baseline is a many-to-one sequence model trained on recent multi-day inputs (wildfire activity and weather), with prediction for next-day personnel and cost.

\noindent\textbf{LLMs.}
We evaluate two major families of large language models under a consistent setup.  
Regular models emphasize throughput and efficiency, including Gemini~2.5~Flash, GPT-4o, GPT-4.1, and GPT-5 (and their mini variants).  
Reasoning models target multi-step decision-making and compositional reasoning, including Gemini~2.5~Pro, GPT-o3, and GPT-o3-mini.  
All models share the same prompts, RAG module, and output schema to ensure comparability, with no fine-tuning applied.

\begin{figure*}[t]
    \centering
    \includegraphics[width=1\textwidth]{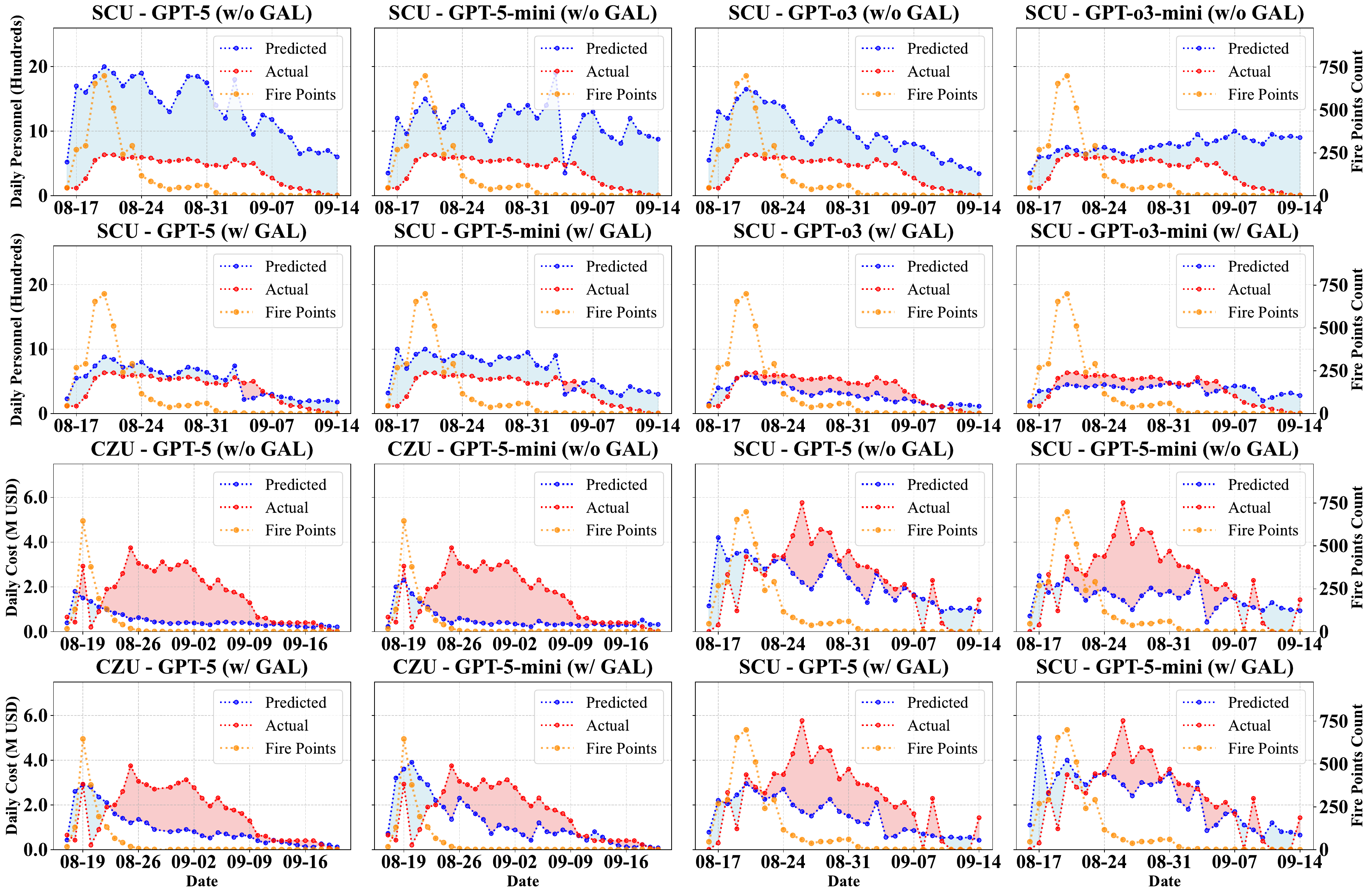}
    \caption{Ablation of the Geospatial Awareness Layer on daily personnel and daily cost forecasting.}
    \label{fig:scu_gal}
\end{figure*}

\subsection{Evaluation Protocol} The task is daily, event-level forecasting: given same-day detections and GAL-derived features, the model predicts (i) required personnel and (ii) daily budget. For a fire of length $N$ with predictions $\hat{y}_i$ and ground truth $y_i$, we report MAE and RMSE. MAE measures the average absolute deviation, whereas RMSE highlights larger errors that show operational under- or over-allocation.

\vspace{-0.3cm}
{\small
\begin{equation*} 
\mathrm{MAE}=\tfrac{1}{N}\!\sum_{i=1}^{N}\!|\hat{y}_i-y_i|,\;\; \mathrm{RMSE}=\sqrt{\tfrac{1}{N}\!\sum_{i=1}^{N}(\hat{y}_i-y_i)^2}. 
\end{equation*}
}

\subsection{Model Performance}
Figure \ref{fig:fig2} summarizes the best cumulative predictive performance achieved by these models across the five held-out wildfires. Tables \ref{tab:personnel}–\ref{tab:cost} report the MAE and RMSE for daily personnel and cost forecasts.

\paragraph{LLMs outperform baselines.}
Across all evaluations, the lowest error in every column is achieved by an LLM.
For daily personnel, GPT-o3-mini, GPT-5-mini, and GPT-4o-mini each lead on different events, while the Physical and LSTM baselines never achieve the top rank.
For daily cost, GPT-o3-mini and GPT-5-mini again dominate most incidents, with Gemini-2.5-Pro and Gemini-2.5-Flash showing competitive results in select cases.
These outcomes highlight the strong generalization of geospatially grounded LLMs.

\paragraph{GAL enhances robustness for complex fires.}
As shown in Figure~\ref{fig:fig2}, complex and spatially fragmented fires such as LNU and August exhibit inherently higher uncertainty, reflected by larger RMSE values.  
Even under these challenging settings, grounded LLMs maintain competitive performance and avoid catastrophic errors, while achieving lower variance across smaller and more localized fires such as El Dorado and SCU.  
This suggests that GAL improves the model’s stability and situational consistency, rather than overfitting to particular spatial configurations.

\paragraph{Compact reasoning models remain competitive.}
Smaller variants such as GPT-o3-mini, GPT-4o-mini, and GPT-5-mini often match or even surpass their larger counterparts across multiple fires.
This counterintuitive pattern arises because the structured and low-entropy inputs provided by GAL reduce linguistic ambiguity and emphasize quantitative reasoning.
Under these stable, unit-normalized conditions, compact models adhere more consistently to schema and numerical cues, achieving higher stability and precision even without the capacity overhead of larger models.

\subsection{Ablation Studies}

We isolate the contribution of each module through two controlled settings: (i) w/o GAL versus w/ GAL, where the former exposes the model only to raw fire detections and coarse daily summaries, and the latter provides the full perception script from the Geospatial Awareness Layer; and (ii) w/o RAG versus w/ RAG, where historical analogs are omitted or appended to the prompt. Prompts, decoding settings, and evaluation protocols are held constant across variants and models.

\begin{figure}[t]
    \centering
    \includegraphics[width=0.49\textwidth]{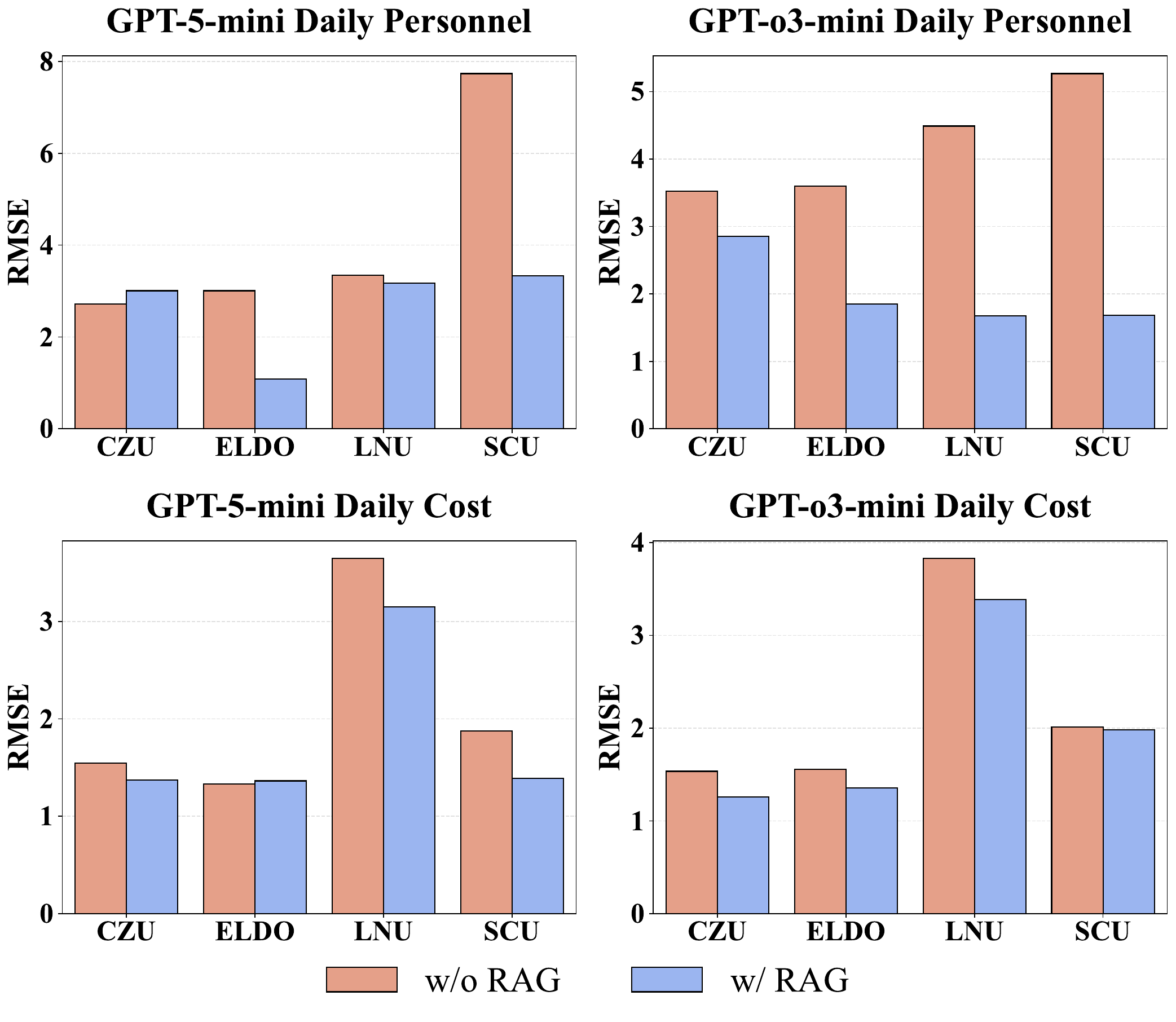}
    \caption{Ablation of the RAG module.}
    \label{fig:rag_personnel}
\end{figure}

\begin{figure*}
    \centering
    \includegraphics[width=1\textwidth]{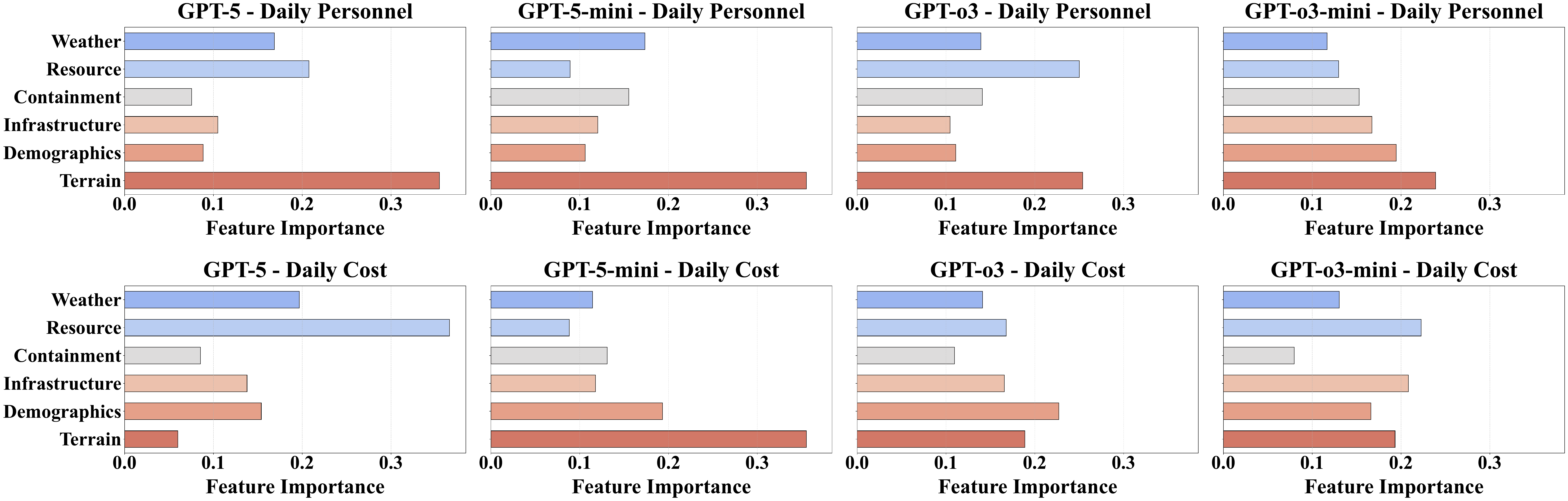}
    \caption{Feature importance analysis of LLM reasoning across four models.}
    \label{fig:feature_importance}
\end{figure*}

Figure~\ref{fig:scu_gal} visualizes the effect of GAL on daily personnel (top two rows) and daily cost (bottom two rows) forecasts across representative wildfire events. In each panel, yellow lines denote fire hotspot counts, red lines show ground-truth records, and blue lines represent model predictions; the shaded areas indicate the magnitude of prediction error. The upper rows compare models without GAL (w/o GAL) and the lower rows show results with GAL (w/ GAL) under identical prompting and evaluation settings. Overall, introducing GAL produces clearer temporal alignment, smoother amplitude transitions, and more realistic decay behavior across both tasks. For SCU personnel, w/o-GAL models substantially overestimate resource levels, maintaining inflated staffing long after the fire subsides. In contrast, w/ GAL predictions follow the actual red curve more closely, with synchronized rise (fall phases and a visibly reduced error region around the incident peak), suggesting better recognition of operational draw-down. For CZU and SCU costs, similar stabilization effects emerge: baseline models tend to underreact early or sustain overly long cost plateaus, whereas w/ GAL variants better capture the inertia of spending.

Figures~\ref{fig:rag_personnel} present RMSE bar charts for daily personnel and cost, comparing w/o RAG (orange) and w/ RAG (blue) across four held-out fires. Lower blue bars indicate improvements from historical retrieval. Overall, RAG reduces error in most cases, though the magnitude varies by model and event. For daily personnel, GPT-o3-mini consistently improves across all fires, with the largest gains on LNU and SCU, while GPT-5-mini achieves clear reductions on ELDO and SCU but shows a slight regression on CZU, likely due to distinctive fire dynamics that make retrieved analogs less relevant. For daily cost, both models achieve steady and robust gains across events. These results suggest that RAG enables models to incorporate cross-event experience to stabilize predictions under sparse or uncertain conditions, enhancing temporal consistency while maintaining geospatial grounding.

\subsection{Failure case analysis}
To illustrate why GAL helps, we present two representative cases in which removing GAL causes systematic behavioral failures.

\begin{figure}[t]
    \centering
    \includegraphics[width=1.02\linewidth]{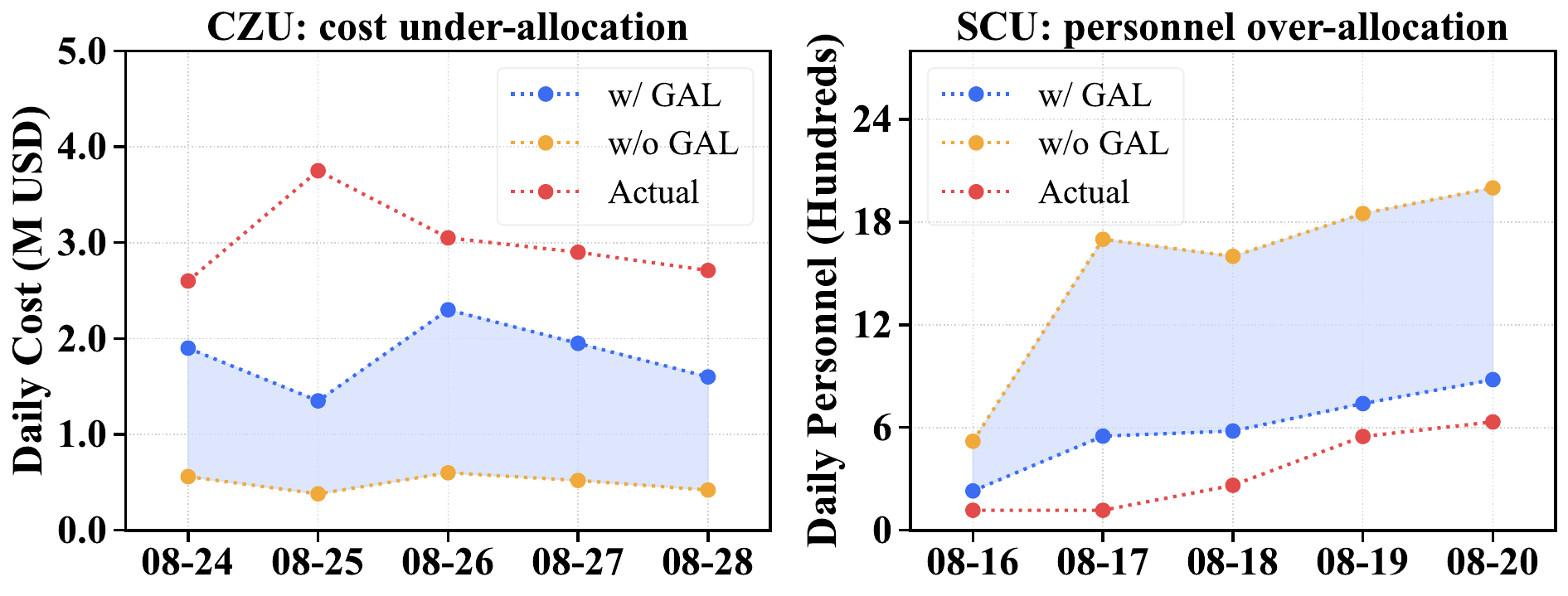}
    \vspace{-4mm}
    \caption{Failure case analysis.}
    \vspace{-3mm}
    \label{fig:failure_cases}
\end{figure}

\noindent \textbf{SCU Lightning Complex: personnel over-allocation.}
Figure~\ref{fig:failure_cases} (left) compares GPT-5 with and without GAL on SCU personnel during 8/16--8/20.
Without GAL, the model relies primarily on FRP and area counts, reasoning: \textit{``Two-cluster fire remains highly energetic\ldots no clear reduction in heat\ldots maintain robust staffing until hotspot density and heat signatures trend down''}, and over-allocates by a factor of 2--3 relative to ground truth.
With GAL, the model explicitly references fuel type, cluster structure, and population exposure: \textit{``Population exposure remains low\ldots Given finite regional resources and limited population density, we stop short of mega-fire levels''}, producing a moderate, resource-constrained trajectory close to observations.
This illustrates a failure mode where FRP-only signals are insufficient to calibrate operationally realistic staffing in low-exposure regions.

\noindent \textbf{CZU August Lightning: cost under-allocation.}
Figure~\ref{fig:failure_cases} (right) compares GPT-5-mini with and without GAL on CZU daily cost during 8/24--8/28.
Without GAL, the model is driven by changes in area and FRP alone, remaining conservative in magnitude and underestimating true cost by more than a factor of four on peak days.
With GAL, the model directly uses terrain and exposure structure---\textit{``fuel continuity is very high in evergreen-dominated terrain\ldots population exposure grew markedly as the large cluster expanded in Santa Cruz''}---and justifies substantial cost increases for aviation, overtime, and structure protection, yielding a trajectory that tracks the high observed costs.

In summary, these cases show that GAL corrects two systematic failures of text-only agents: over-allocation in low-exposure regions and under-allocation in high-risk terrain, yielding operationally meaningful gains.

\subsection{Output Reasoning Analysis}

To better understand what drives the models’ decisions, we train a lightweight post-hoc regressor for each (model, task) pair using the feature labels extracted by the LLMs from the GAL perception script and referenced in their reasoning traces. Features are grouped into six categories: terrain, demographics, infrastructure, containment, resource, and weather. Figure~\ref{fig:feature_importance} reports normalized feature importances. First, for daily personnel prediction, terrain is typically the dominant driver across all four models. Second, for the daily cost task, the influence shifts in a model-dependent way rather than being dominated by a single factor: GPT-5-mini retains terrain as its leading cue, and GPT-o3 variants allocate weight more evenly, with visible contributions. Notably, GPT-5 shows the clearest transition in focus, from terrain in personnel forecasting to resource in cost forecasting, suggesting that larger models may learn to separate operational capacity factors from environmental constraints when estimating expenditures. Additional qualitative examples are provided in \textbf{Appendix~\ref{app:Case_Study}}.


\section{Discussion and Conclusion}
This study bridges the gap between LLMs and the geospatial realities by introducing a Geospatial Awareness Layer that grounds reasoning in structured spatial evidence. From minimal wildfire inputs such as hotspot locations and detection dates, GAL retrieves infrastructure, demographic, terrain, and weather information, fuses them into a compact, unit-locked perception script, and pairs them with historical analogs to guide auditable resource decisions. Empirical evaluations across California fires show that geospatially grounded agents consistently outperform both physical and LSTM baselines in forecasting daily personnel and cost, exhibiting higher accuracy and stronger temporal stability. These findings highlight several key insights. First, structured spatial grounding contributes more to performance than model scale, as smaller reasoning models can match larger ones when provided with normalized and semantically organized inputs. Second, the integration of spatial and temporal cues within GAL enhances robustness, with perception scripts encoding terrain, exposure, and access, while retrieval-augmented analogs supply historical scale priors that moderate overreaction to noisy or incomplete signals. Third, feature attribution reveals that personnel forecasts depend primarily on spread potential and exposure, whereas cost estimation emphasizes accessibility and logistical demand, reflecting real-world operational priorities. Beyond the wildfire context, GAL offers a generalizable interface for other hazards such as floods, hurricanes, and earthquakes, aligning with a broader movement toward foundation-model-driven environmental intelligence built on heterogeneous Earth observations and domain-aware reasoning~\citep{yu2025survey}.

\section{Limitations}

Several limitations warrant further consideration. First, our evaluation focuses on a subset of large 2020 California wildfires. Although GAL is hazard-agnostic and easily extendable, validating its generality across other regions, years, and hazard types (e.g., floods, earthquakes) will require additional data connectors and local calibration. Second, the framework depends on public geospatial datasets (e.g., FIRMS, ACS, HIFLD, NLCD), which may contain detection noise, temporal delays, or resolution mismatches. Our normalization and aggregation steps mitigate these effects, but data quality still constrains absolute accuracy. Third, the feature-importance analysis is post-hoc and descriptive rather than causal. It reveals how GAL-derived factors correlate with model outputs, but further controlled studies are needed to establish causal mechanisms in real operational settings.



\bibliography{main}

\section*{Appendix}

The appendix is organized to support reproducibility, interpretation, and broader comparison. We begin with the full prompt design used in our GAL-based reasoning pipeline, including both system and user prompts, so readers can understand the exact decision interface presented to the models. We then summarize the wildfire events used in the study and provide a qualitative case study that illustrates how grounded reasoning evolves over time for a representative fire. To further contextualize the main results, we include an extended benchmark against additional time-series baselines and LLM variants, followed by token-usage and API-cost statistics that characterize the practical efficiency of different configurations. 

LLM-based assistants were used only for code assistance and minor language polishing.
Our experiments also evaluated the capabilities of several commercial LLMs, including OpenAI GPT and Google Gemini. The total computational and API expenditure for these experiments was approximately \$300.




\appendix

\section{Prompt Design}

\label{app:Prompt_Design}

\begin{promptbox}[SYSTEM PROMPT]          
\begin{lstlisting}[style=promptstyle]
You are a wildfire analysis and resource management expert. You must return ONLY a valid JSON object following the exact schema provided below.

### Global Guidelines
- The task is to estimate TODAY's required daily_personnel and daily_budget.
- reasoning must explain how terrain, weather, fire intensity, population exposure, and resource accessibility shape your judgment, considering both current conditions and previous analysis context.
- daily_personnel is the total integer headcount assigned today (all crews/engines/aviation modules plus command/overhead/support).
- daily_budget is the **new cost incurred today only**, in USD.

### Resource Estimation Principles
- If the fire surges, remember resources are finite-do not assume cost and personnel can scale proportionally.
- When the fire eases, non-suppression needs persist (patrol, mop-up, rehab, logistics); budget and staffing may still be required.
- In "stable" periods, account for cumulative costs and crew fatigue-budgets and crews are not unlimited.
- No detected hotspots $\neq$ full extinguishment; avoid indiscriminate cuts and maintain a prudent baseline.
- Weigh these trade-offs and produce a balanced, defensible recommendation for today's personnel and today's spend. Include any key assumptions and risks.
- **Common pitfall**: after you've committed resources and the fire is "under control" but not yet stable, that actually signals under-resourcing-maintain or increase resources until true stability is confirmed.

### Analysis Approach
- Analyze the fire situation holistically, considering today's conditions and changes from the previous analysis.
- Provide updated estimates for required daily_personnel and daily_budget based on your professional judgment.

### Output Schema (STRICT JSON; no extra keys; no comments)
{
  "analysis_reasoning": {
    "situation_comparison": "<2-3 sentences comparing today vs yesterday>",
    "personnel_reasoning": "<2-3 sentences explaining daily_personnel changes>",
    "budget_reasoning": "<2-3 sentences explaining daily_budget changes>",
    "overall_reasoning": "<2-3 sentences with overall change assessment>"
  },
  "resource_requirements": {
    "daily_personnel": {
      "value": "<integer>",
      "unit": "people"
    },
    "daily_budget": {
      "value": "<integer>",
      "unit": "USD"
    }
  },
  "confidence": {
    "score": "<1-5 integer>"
  },
  "intermediate_indicators": {
    "spread_containment_difficulty": "<minimal|low|moderate|high|critical>",
    "resource_access_deployment": "<minimal|low|moderate|high|critical>",
    "weather_escalation_risk": "<minimal|low|moderate|high|critical>",
    "terrain_operational_complexity": "<minimal|low|moderate|high|critical>",
    "population_exposure_density": "<minimal|low|moderate|high|critical>",
    "fire_station_coverage": "<minimal|low|moderate|high|critical>"
  }
}
\end{lstlisting}
\end{promptbox}

\begin{promptbox}[USER PROMPT]          
\begin{lstlisting}[style=promptstyle]

## Previous Analysis Context
- Previous personnel: <int> people
- Previous daily budget: <int>
- Total cumulative cost: $<int>
- Previous reasoning: <text>

## Cumulative Context
- Total cumulative cost: $<int>
- Total cumulative personnel-days: <int>
- Days since fire start: <int>
- 3-day rolling avg daily cost: $<int>
- 3-day rolling avg daily personnel: <int>
- 7-day rolling avg daily cost: $<int>
- 7-day rolling avg daily personnel: <int>
- Recent cost trend: <increasing|decreasing|stable>
- Recent personnel trend: <increasing|decreasing|stable>

## Fire Intensity Rolling Metrics
- 3-day avg fire points: <float>
- 3-day max fire points: <float>
- 7-day avg fire points: <float>
- 7-day max fire points: <float>
- Current fire points vs historical max: <percent>%
- Global max fire points: <float> (<days> days ago)
- 3-day avg total area: <float> acres
- 7-day avg total area: <float> acres
- Current area vs historical max: <percent>%
- Global max area: <float> acres (<days> days ago)

## Fire Overview vs Yesterday
- Current date: <MM-DD>
- Total Fire Points: <int> (up/down <delta>)
- Num Clusters: <int> (up/down <delta>)
- Total FRP: <float> MW (up/down <delta>, <percent>%)
- Total area: <float> acres (up/down <delta>, <percent>%)
- Max FRP/Brightness: <float> (up/down <delta>)
- Weather conditions: BI=<float>, Tmax/Tmin=<float>, Wind=<float>, FM1=<float>%

## Affected Areas vs Yesterday
- Counties: removed {<names>}; now <int>
- Total Population Affected: <int> (up/down <delta>)
- Fire stations in area: <int> (up/down <delta>)
- Nearest station: <float> mile (up/down <delta>)

## Historical Context (RAG)
- [<FIRE> <DATE>] sim=<float> | Personnel=<float>, Daily_Budget=$<float>
- [repeat for top-K similar cases]

## Cluster Details
- Cluster <ID>:
  fire[points=<int>, frp=<float>, brightness=<float>, area=<float>]
  weather[BI=<float>, tmax=<float>, tmin=<float>, wind=<float>, FM1=<float>]
  location[<County>, <State>, pop=<int>, station_1/2/3=<float> mile]
  terrain: NLCD analysis - land cover, vegetation coverage, fire spread potential
\end{lstlisting}
\end{promptbox}

\section{Wildfire Events and Statistics}
\label{app:Fire_Statistics}

We summarize the fourteen large California wildfire events used in this study in Table~\ref{tab:fire-durations}.  
The dataset covers incidents from late July through early October 2020.  
The top section lists the nine fires used as the historical corpus for model training and RAG retrieval, while the bottom section lists the five events reserved for evaluation.

\begin{table}[t]
  \centering
  \scriptsize
  \setlength{\tabcolsep}{5pt}
  \renewcommand{\arraystretch}{1.05}
  \begin{tabular}{l c c}
    \toprule
    \textbf{Wildfire Event} & \textbf{Start Date (2020)} & \textbf{Duration (days)} \\
    \midrule
    \multicolumn{3}{l}{\textit{Training / Historical Corpus}} \\
    \midrule
    CREEK                 & Sep~5  & 84  \\
    DOLAN                 & Aug~18 & 51  \\
    FORK                  & Sep~8  & 32  \\
    NORTH COMPLEX         & Aug~17 & 83  \\
    RED SALMON COMPLEX    & Jul~28 & 103 \\
    SLATER                & Sep~8  & 61  \\
    SQF COMPLEX           & Aug~24 & 111 \\
    SLINK                 & Aug~29 & 40  \\
    WOODWARD              & Aug~18 & 28  \\
    \midrule
    \multicolumn{3}{l}{\textit{Evaluation / Test Set}} \\
    \midrule
    CZU AUG LIGHTNING     & Aug~17 & 35  \\
    EL DORADO             & Sep~5  & 35  \\
    LNU LIGHTNING COMPLEX & Aug~18 & 29  \\
    SCU LIGHTNING COMPLEX & Aug~16 & 30  \\
    AUGUST COMPLEX        & Aug~17 & 83  \\
    \bottomrule
  \end{tabular}
  \caption{Summary of 2020 California wildfire events.}
  \label{tab:fire-durations}
\end{table}



\section{Case Study: Daily Forecast Trajectory}

\label{app:Case_Study}

\begin{figure*}
    \centering
    \includegraphics[width=1\linewidth]{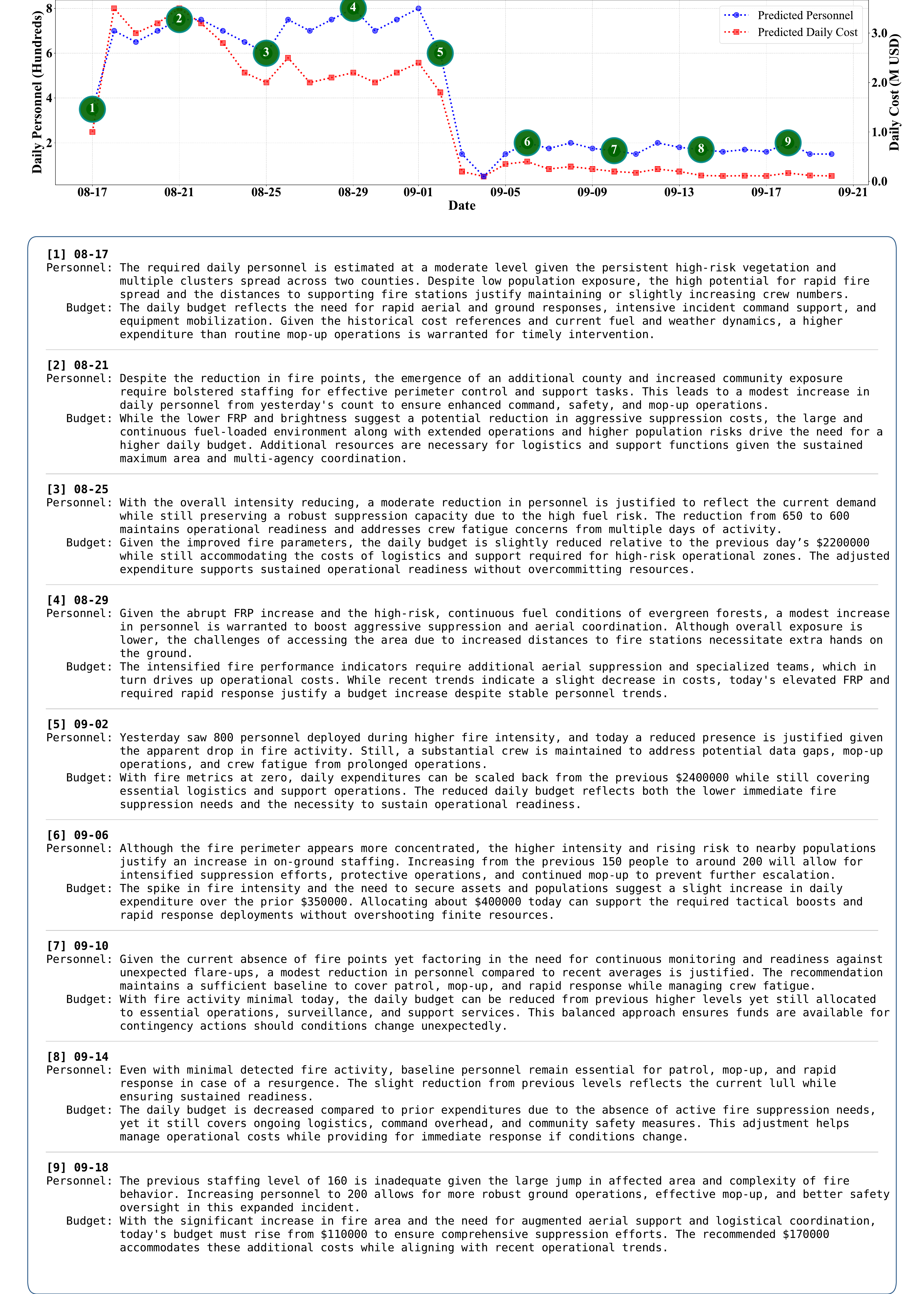}
    \caption{Daily forecast trajectory for the CZU Lightning Complex using GPT-o3-mini with GAL setup.}
    \label{fig:czu_case}
\end{figure*}

Figure~\ref{fig:czu_case} illustrates the daily forecast trajectories of GPT-o3-mini under the GAL-based reasoning setup. The top panel plots the predicted daily personnel and daily cost. The numbered callouts [1]–[9] are the model’s auditable rationales, emitted under our rubric-guided CoT. What follows are concrete behaviors that emerge only because the LLM is reasoning over GAL’s structured context and analog priors. These outputs demonstrate how GAL supports grounded and interpretable reasoning in dynamic disaster contexts.

\textbf{Early phase (08--17 to 08--29).} The model links initial staffing decisions to high vegetation risk, multi-county spread, and limited station coverage, maintaining moderate personnel despite modest FRP levels. When activity shifts toward more populated areas (08--21), the agent increases staffing even as FRP declines, prioritizing exposure and accessibility over fire intensity. This behavior reflects multi-factor reasoning enabled by GAL’s structured context, which aligns spatial and demographic factors on comparable scales.

\textbf{Stabilization and drawdown (08--25 to 09--02).} As fire intensity eases, the model gradually reduces personnel and cost without oscillations. Rationales cite “crew fatigue” and “logistical readiness,” showing that the LLM draws on GAL’s temporal anchors and analog priors rather than reacting to daily noise. The reduction proceeds smoothly, maintaining essential logistics—evidence that the analog-conditioned soft bounds effectively regularize quantitative outputs.

\textbf{Localized escalation (09--06).} When the footprint becomes more spatially concentrated yet closer to populated zones, the model justifies a small resource increase, citing “higher intensity and rising risk to nearby populations.” This response indicates that GAL’s geometric and demographic signals—such as buffer-based proximity and exposure density—guide context-sensitive adjustments that reflect operational priorities rather than purely numerical intensity.

\textbf{Data-sparse periods (09--10, 09--14).} On days with minimal or missing hotspot detections, the model maintains a conservative baseline for patrol and mop-up operations. Its explanations explicitly reference contingency readiness, demonstrating that analog-based priors provide stability and prevent collapse to zero allocations when satellite observations are incomplete. This robustness is crucial for real-world deployment, where sensing delays and cloud cover are common.

\textbf{Re-escalation (09--18).} When the footprint expands again, the model raises both personnel and cost, supported by references to “augmented aerial support” and “logistical coordination.” These justifications align with renewed increases in FRP and spatial extent, showing that the model adapts promptly when environmental signals re-intensify.

\noindent \textbf{Discussion and remark.} These observations provide clear evidence that GAL enhances the reasoning reliability of LLMs in disaster response. 
It enables explicit multi-objective trade-offs: the agent increases staffing when exposure or access risks outweigh intensity, and scales back when trends and logistics allow—supported by GAL’s unified, unit-locked representation. 
The 3/7-day anchors and Incremental mode promote temporal stability without sacrificing responsiveness, while analog-conditioned priors calibrate personnel and cost magnitudes to realistic ranges, preventing overreaction or drift. 
Even under missing signals, the model produces conservative, rationale-grounded baselines instead of collapsing to zero. 
Finally, GAL improves operational explainability, as each chain-of-thought links decisions to verifiable GIS cues such as counties, stations, fuels, and weather.

\section{Extended Benchmark: Time-Series Baselines and Additional LLMs}

To further validate the robustness of our conclusions, we report results on a stricter four-fire subset—CZU August Lightning, El Dorado, LNU Lightning Complex, and SCU Lightning Complex—against an expanded set of competitors across two tasks: daily total personnel and estimated daily cost-to-date.

\paragraph{Non-LLM baselines.}
Beyond the Physical and LSTM baselines reported in the main paper, we evaluate four strong time-series models: DLinear, TimesNet, PatchTST, and TimeLLM. These methods represent a broad spectrum of modern sequence-forecasting approaches, from simple linear decomposition to patch-based transformers and LLM-augmented forecasters. We also include a \textbf{Text-RAG (no GAL)} baseline, which uses the same historical-analog retrieval module as our full system but feeds only unstructured textual summaries of recent FRP and area history to the LLM, without any structured geospatial perception script. This baseline directly tests whether generic retrieval alone—without structured spatial grounding—is sufficient to match GAL-grounded agents. 

\paragraph{Additional LLMs.}
Our main evaluation already spans a diverse set of model families and sizes, including Gemini~2.5~Flash/Pro, GPT-4/4.1/4o/5, and the o3 series. In additional runs, we further evaluate DeepSeek-V3.2, Gemini-3-Pro, GPT-5.1, and a GPT-5 configuration with web-enabled retrieval (GPT-5-web), the last of which approximates a simpler web-search--based RAG baseline without structured geospatial context.

\paragraph{Results.}
As shown in Tables~\ref{tab:personnel_extended}--\ref{tab:cost_extended}, modern sequence models such as DLinear, TimesNet, and PatchTST perform comparably to or worse than the Physical baseline, underscoring the difficulty of temporal extrapolation without spatial context. TimeLLM achieves more competitive results on cost prediction, yet still trails the best GAL-grounded LLMs on most fires. The Text-RAG configuration consistently underperforms full GAL agents, confirming that structured geospatial grounding contributes beyond generic retrieval alone. Among the newly added LLMs, GPT-5-web and Gemini-3-Pro are competitive in select settings, while DeepSeek-V3.2 and GPT-5.1 show moderate performance. Across both tasks and all four fires, GAL-grounded LLM agents consistently match or outperform all baselines, reinforcing that structured geospatial grounding adds value beyond generic retrieval or temporal extrapolation alone.

\begin{table*}[t]
  \centering
  \footnotesize
  \setlength{\tabcolsep}{6pt}
  \renewcommand{\arraystretch}{1}
  \begin{tabular}{c *{8}{S[table-format=1.4]}}
    \toprule
    \multirow{2}{*}{Models}
      & \multicolumn{2}{c}{CZU}
      & \multicolumn{2}{c}{ELDO}
      & \multicolumn{2}{c}{LNU}
      & \multicolumn{2}{c}{SCU}\\
    \cmidrule(lr){2-3}\cmidrule(lr){4-5}\cmidrule(lr){6-7}\cmidrule(lr){8-9}
     & {MAE} & {RMSE} & {MAE} & {RMSE} & {MAE} & {RMSE} & {MAE} & {RMSE} \\
    \midrule
    Text-RAG (no GAL) & 2.8957 & 3.4167 & 1.8742 & 2.1816 & 4.2935 & 4.9984 & 1.6869 & 1.9685 \\
    DLinear       & 5.9859 & 6.3614 & 3.4593 & 3.7735 & 7.3808 & 7.7136 & 3.9831 & 4.4761 \\
    TimesNet      & 4.9538 & 5.7916 & 4.2132 & 5.3438 & 6.2882 & 6.9969 & 5.2743 & 6.9569 \\
    PatchTST      & 3.5742 & 3.9399 & 1.3680 & 1.6111 & 4.7462 & 5.1846 & 2.2235 & 2.4021 \\
    TimeLLM       & 6.7984 & 7.7380 & 1.6207 & 1.9352 & 6.1624 & 6.5885 & 2.8362 & 2.9376 \\
    \midrule
    Gemini\hyp{}2.5\hyp{}Flash & 4.4517 & 5.1462 & 1.0977 & 1.3941 & 4.6493 & 5.3697 & 2.0380 & 2.8660 \\
    GPT\hyp{}4o\hyp{}mini      & 4.0891 & 4.6469 & 1.9234 & 2.1657 & 3.9590 & 4.5367 & {\bfseries 1.0770} & {\bfseries 1.5466} \\
    GPT\hyp{}5\hyp{}mini       & \underline{2.4600} & 3.0067 & \underline{0.9086} & {\bfseries 1.0845} & \underline{2.6693} & \underline{3.1717} & 3.0500 & 3.3324 \\
    GPT\hyp{}5                 & 2.8482 & 3.2180 & 0.9126 & 1.1722 & 3.0859 & 3.7891 & 1.5530 & 1.7762 \\
    GPT\hyp{}5\hyp{}web        & 2.5851 & \underline{2.8872} & {\bfseries 0.8886} & \underline{1.1016} & 3.0172 & 3.6741 & 4.1640 & 4.3954 \\
    GPT\hyp{}5.1               & 4.5651 & 5.1041 & 1.5174 & 1.7251 & 4.0810 & 4.7041 & 1.9830 & 2.2630 \\
    \midrule
    Gemini\hyp{}2.5\hyp{}Pro   & 4.8717 & 5.7561 & 1.9129 & 2.4405 & 4.1486 & 4.9068 & 3.1723 & 4.7006 \\
    Gemini\hyp{}3\hyp{}Pro     & 4.0097 & 4.4446 & 1.2997 & 1.4568 & 3.8729 & 4.3779 & 3.1807 & 4.4384 \\
    DeepSeek\hyp{}V3.2         & 3.5914 & 4.1162 & 1.3285 & 1.6747 & 4.1000 & 4.7346 & 1.9467 & 2.4835 \\
    GPT\hyp{}o3\hyp{}mini      & {\bfseries 2.2826} & {\bfseries 2.8561} & 1.6263 & 1.8525 & {\bfseries 1.2259} & {\bfseries 1.6785} & 1.5093 & 1.6869 \\
    GPT\hyp{}o3                & 3.2583 & 3.7656 & 1.2737 & 1.5457 & 4.0952 & 4.7052 & \underline{1.3927} & \underline{1.6506} \\
    \bottomrule
  \end{tabular}
  \vspace{2pt}
  {\scriptsize \\ Abbrev.: CZU = CZU Aug. Lightning; ELDO = El Dorado; LNU = LNU Lightning; SCU = SCU Lightning.}
  \vspace{-0.2cm}
  \caption{Daily personnel prediction results (MAE/RMSE) on four held-out wildfires.}
  \label{tab:personnel_extended}
\end{table*}

\begin{table*}[t]
  \centering
  \footnotesize
  \setlength{\tabcolsep}{6pt}
  \renewcommand{\arraystretch}{1}
  \begin{tabular}{c *{8}{S[table-format=1.4]}}
    \toprule
    \multirow{2}{*}{Models}
      & \multicolumn{2}{c}{CZU}
      & \multicolumn{2}{c}{ELDO}
      & \multicolumn{2}{c}{LNU}
      & \multicolumn{2}{c}{SCU}\\
    \cmidrule(lr){2-3}\cmidrule(lr){4-5}\cmidrule(lr){6-7}\cmidrule(lr){8-9}
     & {MAE} & {RMSE} & {MAE} & {RMSE} & {MAE} & {RMSE} & {MAE} & {RMSE} \\
    \midrule
    Text-RAG (no GAL) & 1.2283 & 1.3927 & 1.1878 & 1.6621 & 2.5054 & 3.1979 & 1.4997 & 1.8252 \\
    DLinear       & 1.9814 & 2.2738 & 1.3117 & 1.9195 & 3.7606 & 4.4340 & 2.4192 & 2.8928 \\
    TimesNet      & 1.9797 & 2.2729 & 1.3114 & 1.9198 & 3.7585 & 4.4326 & 2.4198 & 2.8917 \\
    PatchTST      & 1.9676 & 2.2618 & 1.2988 & 1.9100 & 3.7502 & 4.4223 & 2.4097 & 2.8812 \\
    TimeLLM       & \underline{0.9768} & {\bfseries 1.2018} & 1.0556 & 1.5233 & 2.7634 & 3.2914 & 1.8859 & 2.0891 \\
    \midrule
    Gemini\hyp{}2.5\hyp{}Flash & 1.4029 & 1.7572 & \underline{0.7971} & 1.3868 & 2.7750 & 3.5688 & 1.6066 & 1.9672 \\
    GPT\hyp{}4o\hyp{}mini      & 1.3520 & 1.6307 & 0.8006 & 1.4770 & 2.9398 & 3.7167 & 1.2701 & 1.5829 \\
    GPT\hyp{}5\hyp{}mini       & 1.0229 & 1.3740 & 0.7973 & \underline{1.3633} & 2.5250 & \underline{3.1504} &  \underline{0.9999} & \underline{1.3888} \\
    GPT\hyp{}5                 & 1.0230 & 1.3284 & 0.8072 & 1.4238 & 2.6781 & 3.3963 & 1.2822 & 1.5542 \\
    GPT\hyp{}5\hyp{}web        & 1.0185 & 1.2800 & {\bfseries 0.7783} & 1.3844 & 2.6985 & 3.4130 & {\bfseries 0.8963} & {\bfseries 1.1402} \\
    GPT\hyp{}5.1               & 1.3679 & 1.6940 & 0.8313 & 1.5050 & 2.9297 & 3.7163 & 1.8093 & 2.1514 \\
    \midrule
    Gemini\hyp{}2.5\hyp{}Pro   & 1.5842 & 2.1840 & 1.3254 & 2.0212 & \underline{2.3426} & 3.3043 & 2.0097 & 2.5964 \\
    Gemini\hyp{}3\hyp{}Pro     & 1.3816 & 1.7301 & 0.8267 & 1.4926 & {\bfseries 2.2283} & {\bfseries 3.0762} & 1.3474 & 1.5935 \\
    DeepSeek\hyp{}V3.2         & 1.2644 & 1.5260 & 0.8518 & 1.4380 & 2.7602 & 3.5321 & 1.4208 & 1.6561 \\
    GPT\hyp{}o3\hyp{}mini      & {\bfseries 0.9530} & \underline{1.2599} & 0.8418 & {\bfseries 1.3535} & 2.7487 & 3.3869 & 1.6596 & 1.9789 \\
    GPT\hyp{}o3                & 1.2494 & 1.5501 & 0.8804 & 1.5253 & 2.9812 & 3.7631 & 1.7996 & 2.1649 \\
    \bottomrule
  \end{tabular}
  \vspace{2pt}
  {\scriptsize \\ Abbrev.: CZU = CZU Aug. Lightning; ELDO = El Dorado; LNU = LNU Lightning; SCU = SCU Lightning.}
  \vspace{-0.2cm}
  \caption{Daily cost prediction results (MAE/RMSE) on four held-out wildfires.}
  \label{tab:cost_extended}
\end{table*}

\section{Token Usage and API Cost per Fire}
\label{app:Token_Cost}

Table~\ref{tab:token_cost} reports the average token usage and estimated API cost per fire for each model evaluated on the four-fire subset (CZU, ELDO, LNU, SCU). These figures reflect the total tokens consumed and dollar cost accumulated across all days of each fire. The results show that the strongest GAL-grounded configurations (e.g., GPT-o3-mini, GPT-4o, GPT-5-mini) achieve competitive predictive performance at modest per-event costs, which is important for real-time or large-scale deployment.

\begin{table*}[t]
  \centering
  \footnotesize
  \setlength{\tabcolsep}{5pt}
  \renewcommand{\arraystretch}{1}
  \begin{tabular}{c
      S[table-format=6.0] S[table-format=1.2]
      S[table-format=6.0] S[table-format=1.2]
      S[table-format=6.0] S[table-format=1.2]
      S[table-format=6.0] S[table-format=1.2]}
    \toprule
    \multirow{2}{*}{Model}
      & \multicolumn{2}{c}{CZU}
      & \multicolumn{2}{c}{ELDO}
      & \multicolumn{2}{c}{LNU}
      & \multicolumn{2}{c}{SCU} \\
    \cmidrule(lr){2-3}\cmidrule(lr){4-5}\cmidrule(lr){6-7}\cmidrule(lr){8-9}
      & {Tokens} & {Cost (\$)}
      & {Tokens} & {Cost (\$)}
      & {Tokens} & {Cost (\$)}
      & {Tokens} & {Cost (\$)} \\
    \midrule
    GPT-4o-mini      &  74244 & 0.02 &  76087 & 0.02 &  72405 & 0.02 &  80827 & 0.02 \\
    GPT-4o           &  74099 & 0.27 &  75350 & 0.28 &  72319 & 0.25 &  80974 & 0.28 \\
    GPT-4.1-mini     &  77512 & 0.05 &  79295 & 0.05 &  75064 & 0.04 &  84225 & 0.05 \\
    GPT-4.1          &  78276 & 0.25 &  79754 & 0.25 &  75445 & 0.22 &  85020 & 0.25 \\
    GPT-5-mini       & 120626 & 0.13 & 122823 & 0.13 & 110436 & 0.11 & 122551 & 0.12 \\
    GPT-5            & 152329 & 0.97 & 153593 & 0.97 & 134258 & 0.79 & 152889 & 0.91 \\
    GPT-5.1          &  80770 & 0.25 &  82097 & 0.25 &  77264 & 0.22 &  86662 & 0.24 \\
    \midrule
    GPT-o3-mini      & 137115 & 0.40 & 139130 & 0.40 & 129254 & 0.36 & 137646 & 0.37 \\
    GPT-o3           &  97130 & 0.40 &  98463 & 0.41 &  89818 & 0.34 & 101459 & 0.39 \\
    \midrule
    Gemini-2.5-Flash &  75903 & 0.06 &  76725 & 0.06 &  71372 & 0.05 &  75656 & 0.06 \\
    Gemini-2.5-Pro   &  74807 & 0.24 &  76364 & 0.25 &  70005 & 0.20 &  77538 & 0.22 \\
    Gemini-3-Pro     &  75192 & 0.68 &  75911 & 0.68 &  70859 & 0.64 &  77861 & 0.70 \\
    DeepSeek-V3.2    & 164132 & 0.06 & 165856 & 0.06 & 146846 & 0.05 & 157016 & 0.06 \\
    \bottomrule
  \end{tabular}
  {\scriptsize \\ Abbrev.: CZU = CZU Aug.\ Lightning; ELDO = El Dorado; LNU = LNU Lightning; SCU = SCU Lightning.}
  \caption{Average token usage and API cost (\$) per fire per model on four held-out wildfires.}
  \label{tab:token_cost}
\end{table*}

\end{document}